\documentclass[letterpaper, 10 pt, conference]{ieeeconf}  
\IEEEoverridecommandlockouts 
\usepackage{amsmath}
\usepackage{amssymb}
\usepackage[caption=false]{subfig}
\usepackage{graphicx} 
\usepackage{float}
\usepackage{leftindex}
\usepackage{url}
\usepackage{fontawesome5} 
\usepackage{pifont}  
\usepackage[noadjust]{cite}
\usepackage[table]{xcolor}
\usepackage[font=small]{caption}
\usepackage{flushend}
\usepackage{slashed}

\newcommand{\cmark}{\ding{51}}
\newcommand{\xmark}{\ding{55}}
\usepackage{xifthen}
\usepackage{xparse}
\usepackage{subdepth}
\usepackage{leftidx}
\usepackage{bm}

\newcommand{\bbm}{\begin{bmatrix}}
\newcommand{\ebm}{\end{bmatrix}}

\DeclareMathAlphabet{\mbf}{OT1}{ptm}{b}{n}
\newcommand{\mbs}[1]{{\bm{#1}}}

\newcommand{\mbsbar}[1]{{\overline{\boldsymbol{#1}}}}
\newcommand{\mbshat}[1]{{\hat{\boldsymbol{#1}}}}
\newcommand{\mbstilde}[1]{{\tilde{\boldsymbol{#1}}}}

\newcommand{\mbsdot}[1]{{\dot {\boldsymbol{#1}}}}

\newcommand{\mbfbar}[1]{{\overline{\mbf{#1}}}}
\newcommand{\mbfhat}[1]{{\hat{\mbf{#1}}}}
\newcommand{\mbftilde}[1]{{\tilde{\mbf{#1}}}}

\newcommand{\mbfdot}[1]{{\dot{\mbf{#1}}}}

\newcommand{\cframe}[1]{{\smash{\protect\underrightarrow{\mathcal{F}}_{#1}}}}

\DeclareMathAlphabet{\mathbfit}{OML}{cmm}{b}{it}
\newcommand{\homo}[1]{{\mathbfit{#1}}}

\newcommand{\mbfh}[1]{{\homo{#1}}}

\newcommand{\pos}[2]{\leftidx{_{#1}}{ \mbf r}{_{#2}}} 
\newcommand{\poshat}[2]{\leftidx{_{#1}}{\mbfhat r}{_{#2}}} 
\newcommand{\posh}[2]{\leftidx{_{#1}}{\mbfh r}{_{#2}}} 

\newcommand{\vel}[3]{\leftidx{_{#1}}{\mbf v}{\IfValueTF{#2}{_{#2#3\hspace{2pt}}}{}}} 
\newcommand{\veltilde}[3]{\leftidx{_{#1}}{\mbftilde v}{\IfValueTF{#2}{_{#2#3\hspace{2pt}}}{}}} 
\newcommand{\velbar}[3]{\leftidx{_{#1}}{\mbfbar v}{\IfValueTF{#2}{_{#2#3\hspace{2pt}}}{}}} 
\newcommand{\velhat}[3]{\leftidx{_{#1}}{\mbfhat v}{\IfValueTF{#2}{_{#2#3\hspace{2pt}}}{}}} 
\newcommand{\veldot}[3]{\leftidx{_{#1}}{\mbfdot v}{\IfValueTF{#2}{_{#2#3\hspace{2pt}}}{}}} 

\newcommand{\acc}[3]{\leftidx{_{#1}}{\mbf a}{\IfValueTF{#2}{_{#2#3\hspace{2pt}}}{}}} 
\newcommand{\acctilde}[3]{\leftidx{_{#1}}{\mbftilde a}{\IfValueTF{#2}{_{#2#3\hspace{2pt}}}{}}} 
\newcommand{\accbar}[3]{\leftidx{_{#1}}{\mbfbar a}{\IfValueTF{#2}{_{#2#3\hspace{2pt}}}{}}} 

\newcommand{\rotvel}[3]{\leftidx{_{#1}}{\mbs \omega}{\IfValueTF{#2}{_{#2#3\hspace{2pt}}}{}}} 
\newcommand{\rotveltilde}[3]{\leftidx{_{#1}}{\mbstilde \omega}{\IfValueTF{#2}{_{#2#3\hspace{2pt}}}{}}} 
\newcommand{\rotvelbar}[3]{\leftidx{_{#1}}{\mbsbar \omega}{\IfValueTF{#2}{_{#2#3\hspace{2pt}}}{}}} 
\newcommand{\rotvelhat}[3]{\leftidx{_{#1}}{\mbshat \omega}{\IfValueTF{#2}{_{#2#3\hspace{2pt}}}{}}} 
\newcommand{\rotveldot}[3]{\leftidx{_{#1}}{\mbsdot \omega}{\IfValueTF{#2}{_{#2#3\hspace{2pt}}}{}}} 

\newcommand{\T}[2]{\leftidx{}{\mbfh T}{_{#1#2\hspace{2pt}}}} 
\newcommand{\q}[3]{\leftidx{_{#3}}{\mbf q}{_{#1#2\hspace{2pt}}}} 
\newcommand{\qtilde}[2]{\leftidx{}{\mbftilde q}{_{#1#2\hspace{2pt}}}} 
\newcommand{\qhat}[2]{\leftidx{}{\mbfhat q}{_{#1#2\hspace{2pt}}}} 

\newcommand{\real}{\mathbb{R}}
\newcommand{\et}{\emph{et al.} }

\title{\LARGE \bf GloPro: Globally-Consistent Uncertainty-Aware \\3D Human Pose Estimation \& Tracking in the Wild}

\author{Simon Schaefer$^{1}$, Dorian F. Henning$^{2}$ and Stefan Leutenegger$^{1,2}$%
\\\tt\small \{simon.k.schaefer, stefan.leutenegger\}@tum.de, d.henning@imperial.ac.uk
\thanks{This research project/publication was supported by TUM AGENDA 2030, funded by the Federal Ministry of Education and Research (BMBF) and the Free State of Bavaria under the Excellence Strategy of the Federal Government and the Länder as well as by the Hightech Agenda Bavaria. Furthermore, it has received financial support from the Munich Data Science Institute (MDSI) through the Seed Fund 2022 project HuMoCap.}%
\thanks{$^{1}$Smart Robotics Lab, Department of Informatics, Technical University of Munich, Germany}%
\thanks{$^{2}$Smart Robotics Lab, Dept.\ of Computing, Imperial College London, UK}%
}

\begin{document}
\maketitle

\begin{abstract}
An accurate and uncertainty-aware 3D human body pose estimation is key to enabling truly safe but efficient human-robot interactions. Current uncertainty-aware methods in 3D human pose estimation are limited to predicting the uncertainty of the body posture, while effectively neglecting the body shape and root pose. In this work, we present GloPro, which to the best of our knowledge the first framework to predict an uncertainty distribution of a 3D body mesh including its shape, pose, and root pose, by efficiently fusing visual clues with a learned motion model. We demonstrate that it vastly outperforms state-of-the-art methods in terms of human trajectory accuracy in a world coordinate system (even in the presence of severe occlusions), yields consistent uncertainty distributions, and can run in real-time. Our code will be released upon acceptance at \url{https://github.com/smartroboticslab/GloPro}.
\end{abstract}

\section{Introduction}
\label{sec:introduction}
Human pose estimation in monocular image sequences is an important cornerstone to understanding and interpreting human behavior in the wild.
Recent research made impressive progress in recovering 3D human meshes with high accuracy, trying to enable common applications such as autonomous driving, home robotics, virtual and augmented reality, and robotic-assisted living.
As humans, we are constantly in touch with, interacting with, or moving around in 3D spaces and objects, and therefore partly or entirely occluded to others.
Furthermore, most aforementioned applications feature a dynamic camera, which significantly increases the problem complexity that we are dealing with.
To incorporate humans into robotic tasks such as path and motion planning, or human-robot interaction, 
an uncertainty-aware representation of the human that is estimated in real-time could potentially drastically increases both the robustness and safety of the robot's actions.
Therefore, the goal of this paper is to tackle the important task of performing human pose estimation and tracking in an uncertainty-aware, causal context.

However, this task is particularly challenging due to several factors.
First, image sequences from dynamic cameras often feature severe occlusions through either dynamic objects, missed or wrong human detections, or scenes where the human or the camera moves in a way such that the human is exiting the field of view of the observing camera.
In cases where there are part occlusions, current mesh regression methods fail, and only a few works have attempted to tackle this challenge \cite{jiang2020multiperson,fieraru2020threedinteract}.
However, those systems fail to integrate temporal context into their predictions, potentially missing out on important details and additional information that would be available in a robotic system.
Secondly, the information in a robotic context is only available in a causal manner, meaning that we only have access to the current and past images.
This complicates the task of motion prediction and infilling as potential end-poses and postures are not known during the first occurrence of the occlusion.
Approaches such as GLAMR~\cite{yuan2022glamr} require that the entire video sequence is known \emph{a priori}, which is in conflict with the premise of robotics/state estimation.

\begin{figure}[t!]
    \centering
    \includegraphics[width=0.48\textwidth]{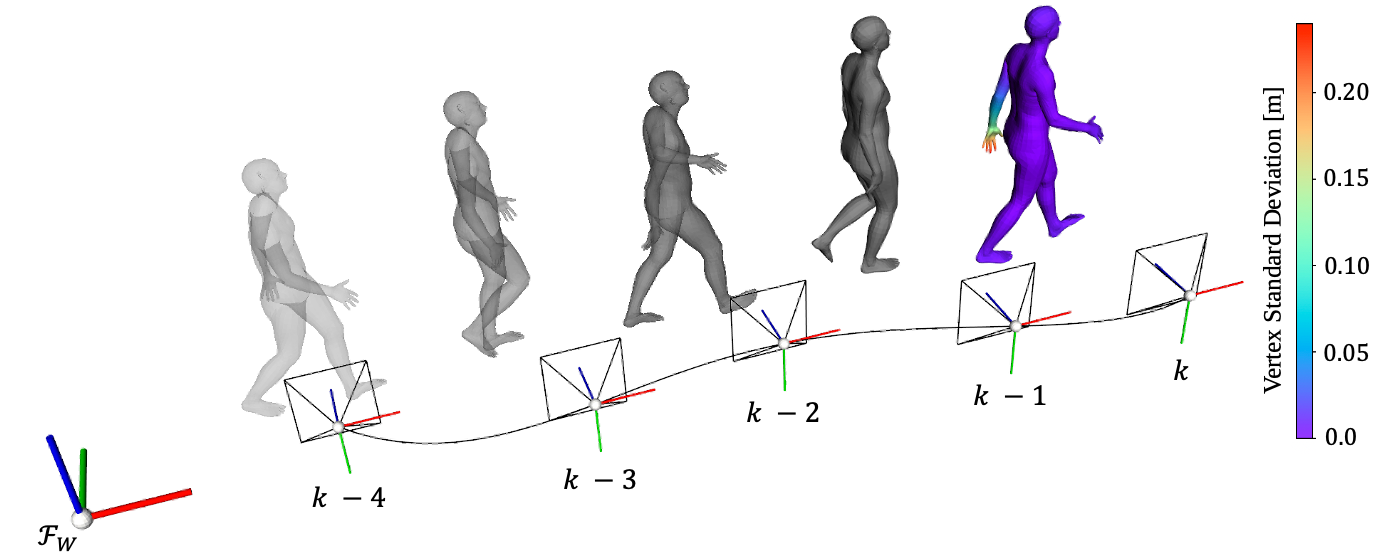}
    \caption{Our method predicts the uncertainty distribution of a human body mesh by fusing visual clues with a motion model. Thus, even in the presence of short-term occlusions, our model is able to predict an accurate body state while stating its uncertainty in regard to the occluded body parts.}
    \label{fig:hero_figure}
\end{figure}

To tackle those challenges, we propose our system \textbf{Glo}bally Consistent \textbf{Pro}babilistic Human Motion Estimation (GloPro), which is a combination of an uncertainty-aware human mesh regressor and a human motion model that predicts the motion based on the joint probability distribution of human mesh parameters in its own, body-centric coordinate frame.
We start by extending a human mesh regressor, first introduced by \cite{kanazawaHMR18}, with a decoder that independent Gaussian distributions for the full body parameter space (mean and variance).
The motion model is a sequence model that accumulates the parameter error states over multiple time steps and produces a forecast about the detected human motion.
Both models can be either independently \mbox{(pre-)trained} or end-to-end, and be used in other contexts (e.g., only the uncertainty-aware human mesh regressor on its own for occlusion-aware human mesh reconstruction).

To summarize, our contributions are the following:
\begin{itemize}
    \item We regress and track the uncertainty distribution of multiple 3D human body meshes parameterized by the SMPL parameters \cite{Loper2015SMPL:Model} from a sequence of RGB images, including the mesh's root translation and orientation. 
    We do not rely on expensive sampling operations or in-the-loop mesh optimization and can achieve real-time performance.
    \item By predicting the body's motion in its own coordinate frame, we disentangle body and camera movement. 
    We show that this improves the model's generalization capabilities and robustness.
    \item We demonstrate the superior performance of our framework in a large variety of scenarios, including occlusions and dynamic scenes.
    While our system shows similar performance to state-of-the-art methods after globally aligning the predicted human mesh to the ground truth, it significantly outperforms those methods in the unaligned case by reducing the reconstruction error by up to $47\%$ and being $20\%$ more run-time and $76\%$ more memory efficient.
\end{itemize}

\section{Related Work}
Since the related work is vast, we will focus on the major aspects of our contribution, the human mesh recovery from a video, the probabilistic modeling, and the frame-to-frame motion prediction.

\subsection{Probabilistic Human Mesh Reconstruction}
Most state-of-the-art approaches to human mesh reconstruction (recovery) assume a fully visible person in a tightly cropped (RGB) image~\cite{kolotouros2019spin,kanazawaHMR18,kocabas2019vibe,iqbal2021kama,Guler2019HoloPose:In-The-Wild,Pavlakos2017Coarse-to-finePose,Pavlakos2018LearningImage,sun2019dsd-satn}.
A few methods try to tackle the case of (self-)occlusions and cropped images where only parts of the person are visible, which is a common problem when using moving cameras or indoor scenes.
These occlusion-aware approaches to human pose estimation show impressive results but are mostly limited to particular settings from where it is hard to generalize.
For single images,~\cite{Kocabas_PARE_2021,fieraru2020threedinteract,zhangoohcvpr20,Rockwell2020,biggs2020multibodies,joo2020eft} are prominent examples.
These methods are mostly trained on artificially occluded images, with either placing photorealistic objects on top of the person, or cropping parts of the image.
Biggs \et~\cite{biggs2020multibodies} extend the human mesh regressor~\cite{kanazawaHMR18} with additional prediction heads to produce multiple hypotheses, while Li and Lee~\cite{Li_2019_CVPR} are generating human posture hypotheses using a Mixture Density Network.
In~\cite{Kocabas_PARE_2021}, the authors use 2D part attention masks to guide the attention of the mesh regressor.
This improves the accuracy but does not supply information about the uncertainty of the predicted mesh.
Sengupta \et \cite{sengupta2021cvpr} use a Gaussian distribution to model the uncertainty of the regressor that estimates the human shape and posture from multiple, unconstrained images.
While this work is an impressive effort and shows how the inclusion of uncertainty measures can improve prediction performance, our work extends this by using causal information and a motion model to produce frame-to-frame predictions.

In~\cite{yuan2022glamr}, they use a generative motion infiller to propose motion samples during long occlusions, before a global optimization is performed.
While both of these methods show impressive results and a large variation in proposed motions, the formulations are non-causal and are very expensive to compute and thus not applicable to the real-time setting we target. 

Another recent work is~\cite{kolotouros2021prohmr}, where the authors embrace the reconstruction ambiguity posed by occluded or cropped 2D image evidence and propose to learn a mapping from the input to a distribution of possible human poses using normalizing flows.
While we appreciate the effort, this approach does not incorporate temporal information into the prediction model, is limited to predicting the body pose uncertainty, and requires computationally expensive sampling to obtain an uncertainty distribution.
Our work, in contrast, combines information from a sequence of images and generates a fully probabilistic reconstruction of human meshes including the root pose and body shape uncertainty.

\subsection{Human Motion Models}
Human motion prediction, synthesis, and infilling were studied extensively in the past.
While there are many approaches that estimate human motion only from 3D key points~\cite{Dabral2017LearningMotion,Hossain2017ExploitingEstimation,Pavllo20183DTraining,ruiz2019human}, we focus on motion models that also utilize a parametric human mesh model.
While earlier works use simple smoothing priors rather than defining an explicit motion model~\cite{Arnab2019}, more recent works are leveraging learned human motion dynamics to improve the accuracy of the human pose estimates~\cite{Kanazawa2019LearningVideo,kocabas2019vibe}.
However, as these methods do not differentiate camera from human body motion it can be doubted whether they can generalize to unseen or dynamic camera motions. 
In contrast, we inherently disentangle these motions by predicting the human body motion independent from the camera motion in a local reference frame.

Rempe \et \cite{rempe2021humor} and Ling \et \cite{ling2020MVAE} both use conditional Variational Autoencoders (cVAEs) to sample realistic human movements from a latent space.
While \cite{ling2020MVAE} focuses on generating realistic motions for a character controller input, HuMoR~\cite{rempe2021humor} uses the cVAE as a prior during a test-time optimization routine.
As impressive as the generated motions are, the realism comes at a cost of complexity; the optimization of HuMoR~\cite{rempe2021humor} takes more than 5 minutes for a 3-second clip.
Some works show the possibility of predicting global translations of the human body from the underlying posture and shape.
Schreiner \et~\cite{schreiner2021globalposition} predict the translations in a global coordinate frame, while Henning \et~\cite{henning2022eccv} estimate them in the body-centric coordinate frame from mesh parameters and use the predictions in a global camera and human bundle adjustment.

In contrast to prior work, we do not require sampling or expensive (test-time) optimization, but only a single forward pass of the previously regressed human mesh parameters.
Furthermore, our method is agnostic to the input format, meaning that we can either supply single images as well as image sequences of arbitrary length and still receive a realistic, causal uncertainty estimate of the mesh parameters and the motion prediction.

\begin{figure*}[ht!]
    \centering
    \includegraphics[width=\textwidth]{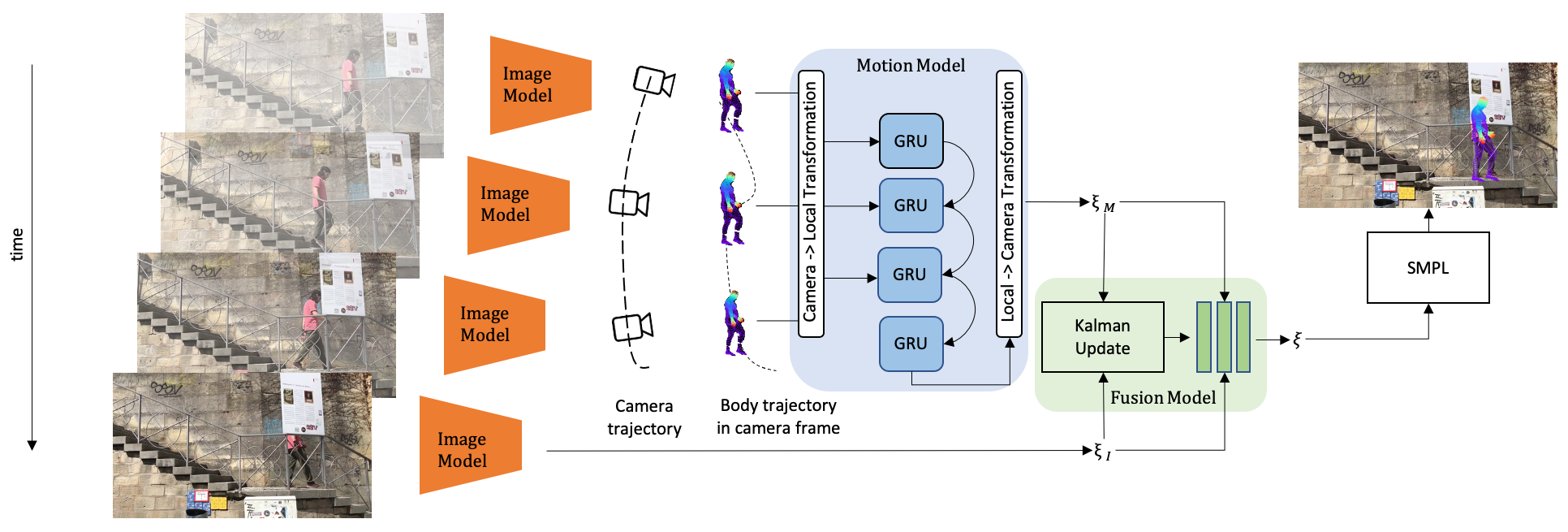}
    \caption{Model Architecture. Our model predicts an image-based prior $\bm{\xi}_I^k$ and a motion-based prior $\bm{\xi}_M^k$ and fuses them to a posterior distribution $\bm{\xi}^k$ over the SMPL parameters~\cite{Loper2015SMPL:Model} afterward. Thereby, given the camera trajectory, the body motion can be estimated independently from the camera motion. The body state distribution can then be propagated to an uncertainty distribution over each vertex of the 3D body mesh.}
    \label{fig:architecture}
\end{figure*}
\section{Prerequisites}
\label{sec:prerequisites}

\subsection{Coordinate Frames, Transformations, and Notation} 
We denote reference coordinate frames as $\cframe{A}$, and points expressed in this reference coordinate frame as $\pos{A}{}$, with their respective homogeneous representation $\posh{A}{}$.
The homogeneous transformation from reference coordinate frame $\cframe{B}$ to $\cframe{A}$ is denoted as $\T{A}{B}$, a function of position vector $\pos{R}{A B}$ and orientation quaternion $\q{A}{B}{}$.
In this work, we use the reference world frame $\cframe{W}$, the camera-centric frame $\cframe{C}$, and the human-centric frame $\cframe{H}$.

It is shown in literature \cite{Zhou2019OnNetworks,kolotouros2019spin,kocabas2019vibe} that many 3D rotation representations are unfavorable for learning tasks, as they suffer from singularities.
Due to the wide variety of literature about the use of quaternions in the context of rotation uncertainty and the straightforward definition of the rotation error state, the body posture parameters are given as quaternions in  vector notation $\q{}{}{} = [\mbf a^T, b]^T$, where $\mbf a$ denotes the imaginary part, and $b$ the real component.
This allows us to define the rotation error state between a rotation $\q{}{}{}$ and its measurement $\qtilde{}{}$ as:
\begin{align}
\delta \q{}{}{} = \qtilde{}{} \otimes \q{}{}{}^{-1},
\end{align}
\noindent
such that the rotation error $\mbf{e}_{\text{q}}$ can be expressed as: 
\begin{equation}
    \mbf{e}_{\text{q}} = \delta \q{1:3}{}{}.
\end{equation}

In the following, we denote the normally distributed random (vector) variable $\mbf{x} \sim \mathcal{N}(\mbfhat{x}, \bm{\Sigma}_x)$ with mean $\mbfhat{x}$ and covariance matrix $\bm\Sigma_x$. Furthermore, the function $\text{vec}(\cdot)$ vectorizes a matrix, i.e.\ for a sample matrix $\mbf{M} = \begin{bmatrix} a & b\\ c & d \end{bmatrix}$, the vectorization is $\text{vec}(\mbf{M}) = \left[a, b, c, d \right]^T$.

\subsection{Human Mesh Representation}
To represent the human body, we adapt the SMPL model~\cite{Loper2015SMPL:Model}, and parameterize the posture vector $\bm{\theta}$ as 23 Hamilton Quaternions.
The body shape is parameterized as $\bm{\beta} \in \real^{10}$.
The SMPL model supplies a function ${_H}\mathcal{S}(\bm{\beta}, \bm{\theta}) = {_H}\mbf{V} = \left[ {_H}\mbf{v}_1, \dots, {_H}\mbf{v}_N \right]$, representing a mesh with $N = 6890$ vertices in the human reference frame.
The body vertices expressed in the camera frame are given as:
\begin{equation}
    {_C}\mbf{V} = \left[ \T{C}{H} {_H}\mbfh{V} \right]_{1:3} =: {_C}\mathcal{S}(\bm{\xi}),
\end{equation}
\noindent
with ${\bm{\xi} = \left[ \bm{\beta}^T, \bm{\theta}^T, \pos{C}{C H}^T, \q{C}{H}{}^T \right]}$ the human state vector -- that we seek to estimate -- consisting of shape $\bm\beta$, posture $\bm\theta$, position in camera coordinates $\pos{C}{C H}$, and orientation relative to the camera frame $\q{C}{H}{}$.


\section{Problem Formulation}
\label{sec:problem_formulation}
Given a sequence of $N$ images $(\mathcal{I}_{k-N}, ..., \mathcal{I}_k)$ with known intrinsic parameters and camera poses $\T{W}{C_k}$, we aim to regress a plausible uncertainty distribution over the 3D mesh of a human body at time-step $k$.
Due to the highly constrained structure of the human body, the individual vertex distributions would be highly correlated.
As a consequence, predicting the full joint distribution of all vertices, similar to~\cite{zhang2021mojo} is not practical: even with the simplifying assumption of a Gaussian distribution, it would require predicting the full covariance matrix and therefore be infeasible to compute, given a large number of 3D vertices in typical human body mesh representations.
Therefore, we instead regress the distribution over the SMPL~\cite{Loper2015SMPL:Model} parameters, which effectively disentangle the body shape, posture, and root position.
For simplicity, we assume that all distributions are uni-modal, approximated by a Gaussian with mean $\bm{\mu}_\xi$ and covariance matrix $\bm{\Sigma}_\xi$:
\begin{align}
    \bm{\xi} \sim \mathcal{N}(\hat{\bm{\xi}}, \bm{\Sigma}_\xi).
\end{align}

\begin{figure}[!h]
    \centering
    \includegraphics[width=0.41\textwidth]{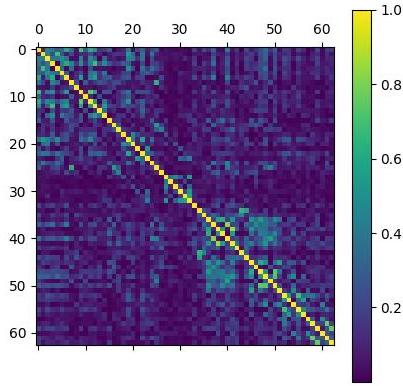}
    \caption{Absolute values of SMPL body posture correlations over AMASS dataset~\cite{AMASS:2019}. The hand postures are neglected as they are assumed to be zero in the dataset. As the non-diagonal correlation terms are dominated by the diagonal terms, the body posture co-variances can be safely neglected.
    }
    \label{fig:corr_amass}
\end{figure}

Due to the disentangled nature of the SMPL~\cite{Loper2015SMPL:Model} body representation, we further chose to ignore the correlations for simplicity. We motivate this choice as shown qualitatively in Fig.~\ref{fig:corr_amass}, which shows the empirical correlation matrix of the SMPL parameters across the whole AMASS dataset~\cite{AMASS:2019}~-- revealing a diagonally-dominant structure. This choice drastically reduces the number of parameters to predict and thus simplifies the overall learning problem.

\section{Approach}
\label{sec:approach}
GloPro leverages information from a sequence of images by explicitly fusing image and motion-based predictions.
In contrast to an implicit fusion of sequential visual clues such as in~\cite{kocabas2019vibe}, our method allows pre-training of the motion-based prediction model that is \emph{independent} from the camera trajectory and is more interpretable. 
With $\bm{\xi}_I^k$ being the single-image-based prediction, $\bm{\xi}_M^k$ the sequence-based prediction, the posterior distribution of $\bf{\xi}^k$ is estimated by a learned fusion model $f_F$:
\begin{align}
    \bm{\xi}^k = f_F(\bm{\xi}_I^k, \bm{\xi}_M^k).
\end{align}

As shown in Fig.~\ref{fig:architecture} that depicts the overall approach, the distribution of the 3D body mesh vertices, ${_C}\mbf{V}$, is then computed via linear error propagation with respect to the SMPL model, using the Jacobian of the vertices with respect to the human state vector, $\mbf{J}_V$:
\begin{align}
    \mbf{J}_V &= \frac{\partial \text{vec}({_C}\mbf{V})}{\partial \bm{\xi}}, \\
    \text{vec}({_C}\mbf{V}) &\sim \mathcal{N}(\text{vec}({_C}\mathcal{S}(\hat{\bm{\xi}})), \mbf{J}_V \bm{\Sigma}_\xi \mbf{J}_V^T).
    \label{eq:vertex_distribution_propagation}
\end{align}

The resulting estimated body mesh can be transformed from the camera to the world frame using the known camera poses. 
This allows for simple but robust tracking independent from the camera motion.

\subsection{Model Architecture}
\textbf{Image Model}
The image model closely follows the architecture of \cite{kanazawaHMR18}.
The model extracts image features from the input image using a Resnet-50~\cite{he2016resnet50} backbone.
Thereby, in order to improve robustness, the focal distortion is removed from the input image by re-projecting it to a standardized virtual camera, similar to~\cite{henning2020hpe3d,yu2020pcls}.
The image features are then combined with the VPoser mean~\cite{pavlaakos2019vposer} to iteratively refine the body state estimate.
An additional MLP is attached to the network's head to regress the logarithmic body parameter variances, based on the refined body state, its difference to the VPoser mean pose, as well as the extracted image backbone features.

\textbf{Motion Model}
Although the body is observed in the camera frame, its motion is independent of any camera movement.
In order for a human motion model to be able to generalize well in the presence of a dynamic camera movement, both motions have to be disentangled. 
Therefore, from the estimated transformation between camera and body $\T{C_k}{B_k}$, we introduce a body frame $\cframe{B_k}$ posed at the estimated body pose at every timestamp.
By leveraging the camera poses $\T{W_{k\text{-}j}}{C_{k\text{-}j}}$, which can e.g.\ be estimated from a SLAM system~\cite{leutenegger2022okvis2,Mur-Artal2015ORB-SLAM:System}, we transform the input body states from the camera frame to relative transformations between the human frames $H_{k\text{-}j}$ and $H_{k\text{-}1}$, i.e.
\begin{align}
    \T{W}{H_{k\text{-}j}} &= \T{W}{C_{k\text{-}j}} \T{C_{k\text{-}j}}{H_{k\text{-}j}}, \\
    \T{H_{k\text{-}1}}{H_{k\text{-}j}} &= \T{W}{H_{k\text{-}1}}^{-1} \T{W}{H_{k\text{-}j}}.
    \label{eq:body2bodytrafo}
\end{align}

Given a sequence of $M$ body states expressed as relative frame transformations in the body frame $\cframe{H_{k\text{-}1}}$ the motion model predicts the body state at time $k$ relative to $\cframe{H_{k\text{-}1}}$.
Therefore, the image-based predicted root-pose distributions 
${\bm{\gamma}} = \left[\pos{C_j}{C_j H_j}^T, \q{C_j}{H_j}{}^T \right]$ are propagated from their respective camera frame $\cframe{C_j}$ to the body reference frame $\cframe{H_{k\text{-}1}}$.
With $\mbf{J}_{CH}$ being the Jacobian of the root-pose in the camera frame with respect to the relative transformation obtained by Eq.~\ref{eq:body2bodytrafo}, the root-pose ${\bm{\tau}_{k-j}}$ in the body reference frame $\cframe{H_{k\text{-}1}}$ can be obtained as:
\begin{equation}
    \bm{\tau}_{k-j} \sim \mathcal{N}\left(\left[\poshat{H_{k\text{-}1}}{H_{k\text{-}1}H_{k\text{-}j}}^T,\qhat{H_{k\text{-}1}}{H_{k\text{-}j}}^T \right], \mbf{J}_{CH} \bm{\Sigma}_{ {\bm{\gamma}}} \mbf{J}_{CH}^T\right),
\end{equation}
where $\pos{H_{k\text{-}1}}{H_{k\text{-}1}H_{k\text{-}j}}$ and $\q{H_{k\text{-}1}}{H_{k\text{-}j}}{}$ correspond to the translational and rotational parts of $\T{H_{k\text{-}1}}{H_{k\text{-}j}}$.
While we omit the math here for brevity, the motion model's output pose distribution is analogously transformed back to the camera frame $\cframe{C_k}$.
This is necessary to fuse the motion and image model's estimates of the current state in the next step.

Inspired by the success of~\cite{kocabas2019vibe} in sequence-based human motion prediction, a GRU is used, followed by an MLP decoder for predicting the output body state distribution.

\textbf{Fusion Model}
The a priori distributions estimated from the image $\bm{\xi}_I^k$ and motion $\bm{\xi}_M^k$ are merged to a single posterior distribution.
Thereby, the model first fuses the prior distributions using a Kalman filter and then adds a learned state-dependent residual to the mean of the Kalman posterior distribution.
For predicting the residual term an MLP is used.
We empirically found that this architecture delivers the best predictive performance, compared to a linear Kalman update or a purely learned approach.

\subsection{Training Objective}
Given the ground-truth body state $\bm\xi^{\mathrm{gt}}$ in the posterior distribution of $\bm{\xi}$, i.e.\ $p_{\xi}(\bm\xi)$, our model can be supervised by the negative log-likelihood loss:
\begin{align}
    \nonumber
    &\mathcal{L_{\mathrm{NLL}}} = - \log p_{\bm{\xi}}(\bm\xi^\mathrm{gt}) \\
    &= (\hat{\bm{\xi}}-\bm{\xi}^{gt})^T \bm{\Sigma}_{\xi}^{-1} (\hat{\bm{\xi}}-\bm{\xi}^{gt})
    - \log \det(\bm{\Sigma}_\xi) + \slashed{c},
\end{align}
with constant $\slashed{c}$.
To support the limited availability of 3D annotated ground truth data, most methods~\cite{kanazawaHMR18,kolotouros2019spin,kocabas2019vibe} supervise the 3D joint positions with a 2D reprojection loss in the image plane via a (weak) perspective projection function.
The ground truth 2D joint position is obtained from the OpenPose model~\cite{Cao2017RealtimeFields} that computes a probability density $p_{H^\text{gt}_i}$ for each joint $i$ over the input image.
The ground truth 2D joint position $\mbf{h}^\text{gt}_i$ then is the maximum likelihood of ${p_{H^\text{gt}_i}}$.
Using the probability density ${p_{H^\text{gt}_i}}$, the re-projection error can be reformulated probabilistically.
The predicted distribution of $\bm{\xi}$ is propagated to the 2D joint space by first computing the distribution of ${_C}\mbf{L}$ concatenating the 3D joints $\mbf{l}_i$, and then projecting to a $\mbf{H}$, the concatenation of the 2D joints $\mbf{h}_i$, using the perspective projection function $\mbs{\pi}(\cdot)$.
With $\mbf{J}_L$ being the Jacobian of the 3D body joints with respect to the human state vector, and $\mbf{J}_\pi$ the well-known Jacobian of the perspective projection function taken from~\cite{Furgale2011ExtensionsVisualOdometry}:
\begin{align}
    \text{vec}({_C}\mbf{L}) &\sim \mathcal{N}(\text{vec}({_C}\mathcal{S}(\hat{\bm{\xi}})), \mbf{J}_L \bm{\Sigma}_\xi \mbf{J}_L^T) \\
    \text{vec}(\mbf{H}) &\sim \mathcal{N}(\text{vec}(\mbs{\pi}({_C}\hat{\mbf{L}})), \mbf{J}_\pi \bm{\Sigma}_L \mbf{J}_\pi^T).
\end{align}

Then, the probabilistic reprojection loss for joint $i$ is the Kullback-Leibler divergence between its corresponding probability distribution ${p_{H^\text{gt}_i}}$ and the predicted distribution ${p_{H_i}}$ over the 2D joint space:
\begin{equation}
    \mathcal{L}_{\mathrm{KL}} = \sum_i KL({p_{H_i}}, {p_{H^\text{gt}_i}}).
\end{equation}

To stabilize the model training, we add the re-projection term based on mean $\hat{\bm{h}_i}$ of the 2D joint distribution and the OpenPose ground truth 2D joint position $\mbf{h}^\text{gt}_i$: 
\begin{equation}
    \mathcal{L}_{\mathrm{RP}} = \sum_i \|\mbfhat{h}_i - \mbf{h}^\text{gt}_i\|^2.
\end{equation}

Lastly, similar to \cite{henning2022eccv}, we regularize the body shape parameters using the L2-norm.
\begin{equation}
    \mathcal{L}_\mathrm{\beta} = ||\hat{\bm{\beta}}||^2.
\end{equation}

In conclusion, the final training objective becomes, applied to frame $k$: 
\begin{equation}
    \mathcal{L} = \mathcal{L}_{\mathrm{NLL}} + \lambda_{KL}\mathcal{L}_{\mathrm{KL}} + \lambda_\mathrm{RP}\mathcal{L}_{\mathrm{RP}} + \lambda_{\beta}\mathcal{L}_{\mathrm{\beta}}.
    \label{eq:loss_full}
\end{equation}

\section{Experiments \& Results}
\label{sec:experiments}
\begin{table*}[htb!]
    \caption{Comparison with several methods for human pose estimation and tracking on the 3DPW dataset \cite{vonMarcard2018}.
    The best result is marked in bold.
    GloPro outperforms the baseline methods by a large margin in the unaligned accuracy metrics {G-MPJPE} and {G-PVE}.
    While most other methods show similar performance on the {PA-MPJPE} metric, one has to note, that \textbf{in the wild Procrustes alignment is not possible} and that this error metric only assesses the shape and posture prediction accuracy, while ignoring global consistency and relative translation and orientation between the human and camera frame.
    }
    \label{tab:performance}
    \centering
    \normalsize
    \begin{tabular}{l|c|c|c||c|c|c|c}
    & Input & Root & Single & G-MPJPE & PA-MPJPE & G-PVE & G-Accel \\
    & & Pose & Stage & [mm] $\downarrow$ & [mm] $\downarrow$ & [mm] $\downarrow$ & [mm/$s^2$] $\downarrow$ \\
    \hline
    HMR \cite{kanazawaHMR18} & Image & \cmark & \cmark & 231.9 & 73.6 & 253.1 & 228.9 \\
    ProHMR \cite{kolotouros2021prohmr} & Image & \cmark & \cmark & - & 70.41 & - & - \\
    PARE \cite{Kocabas_PARE_2021} & Image & \cmark & \cmark & - & \textbf{44.08} & - & - \\
    VIBE \cite{kocabas2019vibe} & Video & \xmark & \cmark & - & 58.97 & - & - \\
    HuMoR \cite{rempe2021humor} & Video & \cmark & \xmark & 211.62 & 48.89 & 225.58 & \textbf{6.0} \\
    \hline
    Ours (No World) & Video & \cmark & \cmark & 175.29 & 50.24 & 190.31 & 223.0 \\    
    Ours (No Cov.) & Video & \cmark & \cmark & 138.72 & 51.1 & 161.81 & 179.07 \\
    \hline
    \textbf{Ours} & Video & \cmark & \cmark & \textbf{114.48} & 48.9 & \textbf{126.28} & 151.06 \\
    \hline
    \end{tabular}
\end{table*}

All experiments within this work have been conducted using the PyTorch framework \cite{Paszke2017AutomaticPyTorch} on a single NVIDIA RTX A4000 GPU. 
We used the 3DPW dataset \cite{vonMarcard2018} for both training and evaluation, as it is the only 3D human pose dataset including heavy occlusion, an outdoor environment, and dynamic camera motion.
It contains about 60 motion sequences (36 for training, 24 for testing) with 3D annotations and camera trajectory.
For training, we used the Adam optimizer \cite{Kingma2015Adam:Optimization} with a learning 0.001 and a batch size of 24, a sequence size of $M = 4$ for the motion model as well as $\lambda_{KL} = \lambda_{RP} = 1.0$ and $\lambda_\beta = 0.001$, which have been empirically found to yield the best model performance.

\subsection{3D Mesh Regression}
In line with~\cite{yuan2022glamr}, we use the un-aligned mean per joint position error (G-MPJPE), the Procrustes-aligned MPJPE (PA-MPJPE), and the un-aligned per vertex error (G-PVE) to evaluate the accuracy of the 3D posed body mesh in a global frame.
To further evaluate the temporal consistency of the predicted body motion, we report the 3D joint acceleration in the global frame (G-Accel).
Note, however, that Procrustes alignment is not possible for true ``in-the-wild'' data, and that this error metric only assesses the human shape and posture accuracy, while ignoring the relative 6D pose between the camera and human frame.
To ensure comparability, we trained all baseline methods on the 3DPW dataset, similar to our method.
Next to several state-of-the-art methods for both image- and video-based 3D human pose estimation, we ablate the proposed disentanglement of the camera and body motion (\textit{Ours (No World)}), and the predictive benefit of our probabilistic formulation in comparison to a fully deterministic method with otherwise similar architecture (\textit{Ours (No Cov)}).
As shown in Table~\ref{tab:performance}, we consistently outperform our baseline methods, but also state-of-the-art methods in terms of the mean 3D body pose estimation.
Notably, our model is able to outperform even multi-step models such as HuMoR \cite{rempe2021humor} by a large margin in terms of predicting a globally consistent body state. 

\subsection{Uncertainty Consistency Analysis}
\begin{figure}[ht!]
    \centering
    \includegraphics[width=0.48\textwidth]{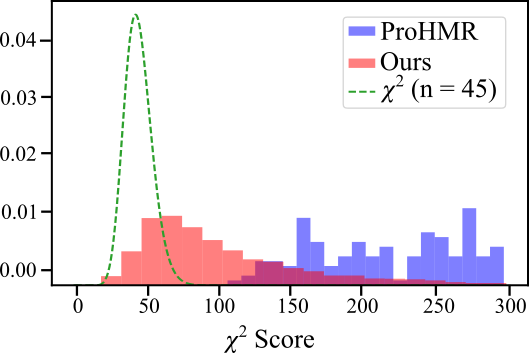}
    \caption{$\chi^2$ consistency analysis of the predicted uncertainty distribution of the 45 SMPL 3D joints \cite{Loper2015SMPL:Model} against the theoretical $\chi^2$ distribution, based on the 3DPW test set \cite{vonMarcard2018}. While our distribution is slightly overconfident, it still can model the theoretical $\chi^2$ distribution much better than our baseline method.}
    \label{fig:chi2_comparison}
\end{figure}
For any uncertainty quantification method, the estimated covariance should roughly model the error between the estimated mean and ground truth.
In order to analyze the consistency of the predicted uncertainty distribution, we, therefore, compute statistics of the $\chi^2$ error.
Ideally, the $\chi^2$ error should thereby follow the theoretical $\chi^2$ density function $p_{\chi^2}$.
However, $p_{\chi^2}$ is not well defined for large degrees of freedom. Therefore, instead of evaluating the consistency of the predicted vertex distribution $\mbf{V}$, we analyze the uncertainty distribution of the predicted 3D joints $\mbf{l}$, which is obtained by propagating the predicted distribution over the body parameters through the SMPL model \cite{Loper2015SMPL:Model}, i.e.\ similar to  Eqn.~(\ref{eq:vertex_distribution_propagation}):
\begin{align}
    {\chi^2} = (\hat{\mbf{l}} - \mbf{l}^{\mathrm{gt}})^T \bm{\Sigma}_l^{-1} (\hat{\mbf{l}} - \mbf{l}^{\mathrm{gt}}).
\end{align}

While several methods predict a probability distribution over the body parameter space, as previously described, we choose ProHMR \cite{kolotouros2021prohmr} as a baseline, as it is, to the best of our knowledge, the only recent method providing open-source accessible code.
Fig.~\ref{fig:chi2_comparison} compares the theoretical $\chi^2$ distribution with the distributions predicted by our method and ProHMR \cite{kolotouros2021prohmr}.
As ProHMR \cite{kolotouros2021prohmr} is a sampling-based method, we statistically approximate the body pose distribution from 4096 samples.
While both predictions are over-confident, the predicted distribution of our method not only provides a more consistent uncertainty estimate but is also more continuous, most likely as it directly regresses the distribution parameters instead of sampling.

\subsection{Runtime Analysis}
\begin{figure}[htb!]
    \centering
    \includegraphics[width=0.48\textwidth]{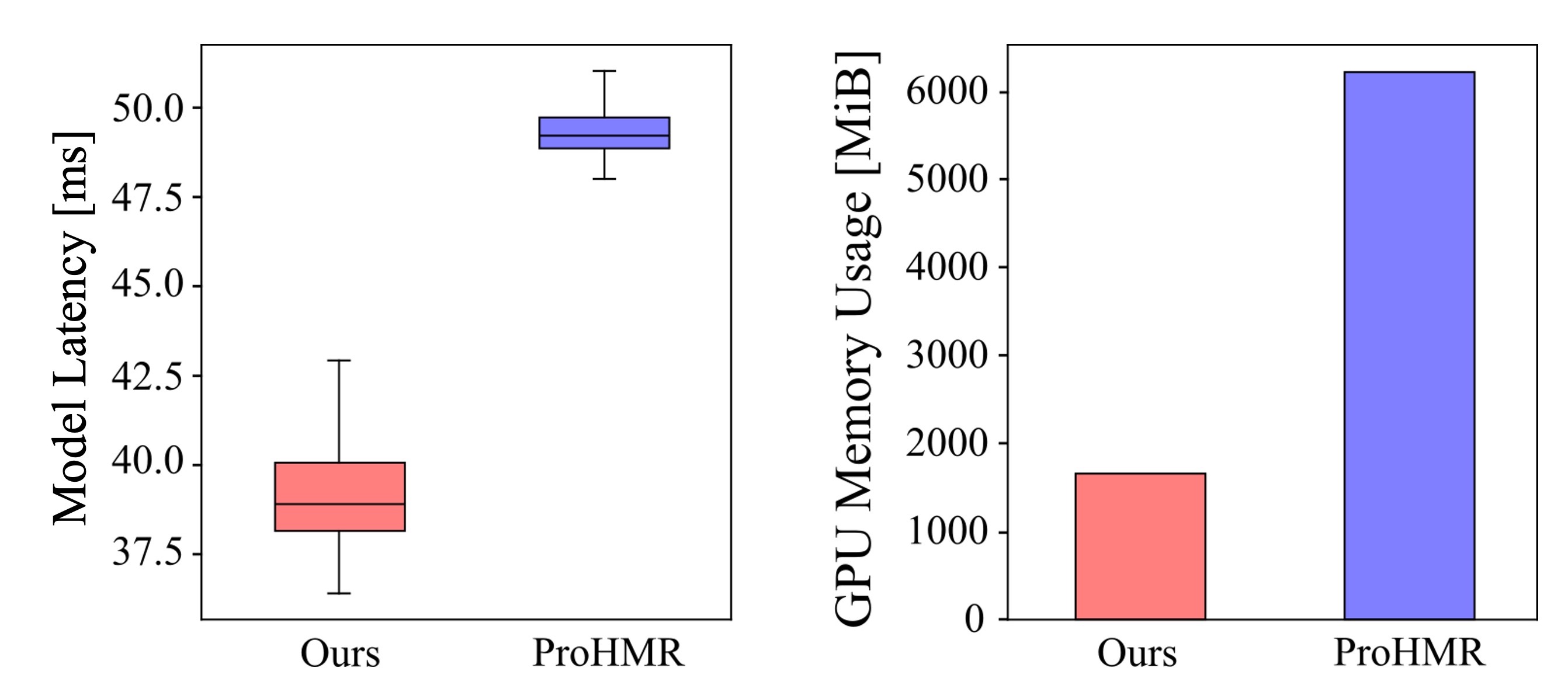}
    \caption{Model latency and memory complexity comparison between our method and ProHMR  \cite{kolotouros2021prohmr}. As shown, our method is much more run time and memory efficient.}
    \label{fig:runtime}
\end{figure}
Fig.~\ref{fig:runtime} shows the run-time and memory complexity of our method averaged over all sequences of the 3DPW test dataset~\cite{vonMarcard2018}. As our method is single-stage, as well as directly regresses the uncertainty distribution without the need for costly sampling, it can run in real-time (20 Hz) and is 4 times more memory efficient than ProHMR \cite{kolotouros2021prohmr}.

\subsection{Qualitative Results}

\begin{figure*}[htb!]
\centering
{\includegraphics[width=0.20\textwidth]{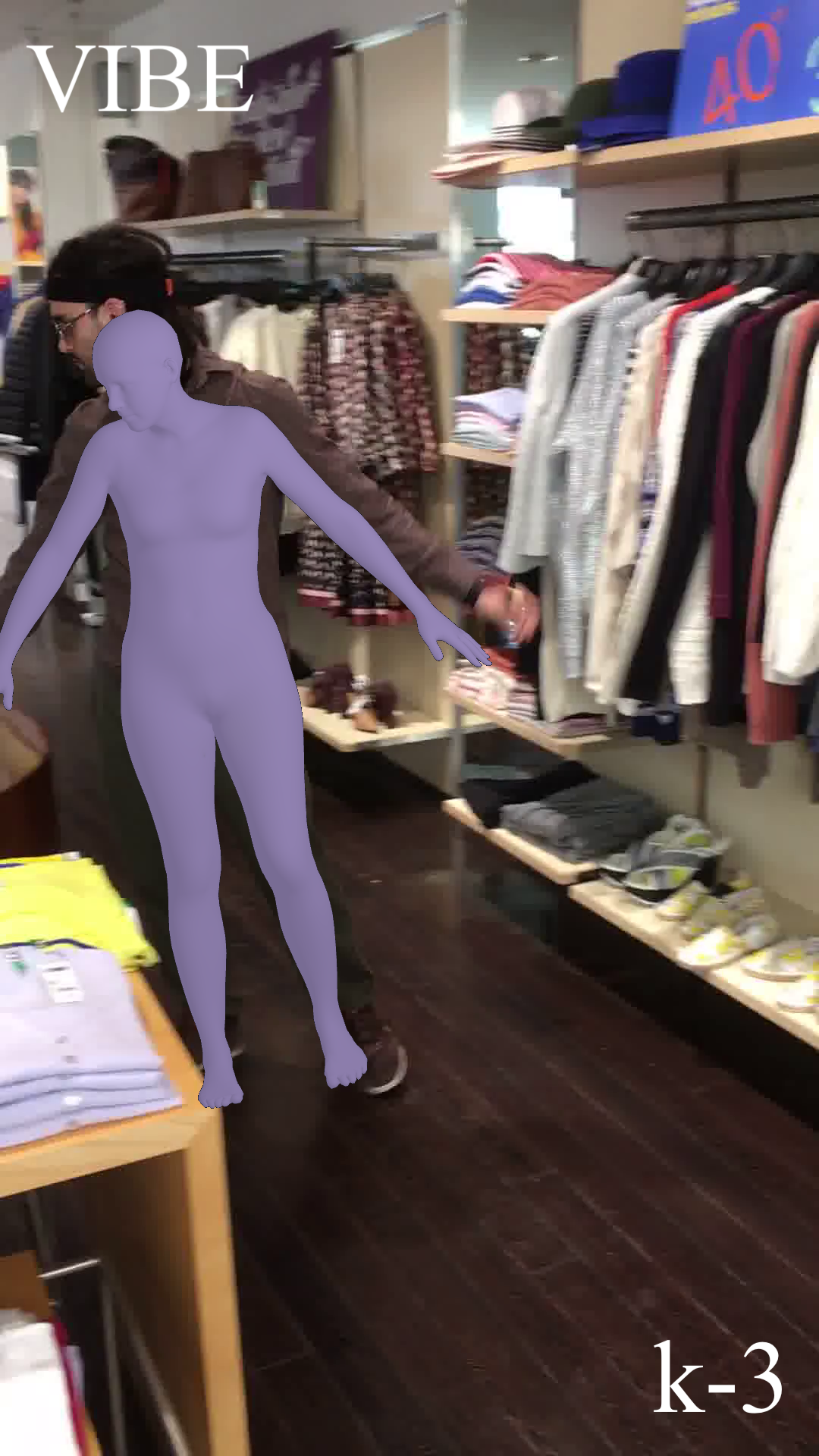}}
\quad
{\includegraphics[width=0.20\textwidth]{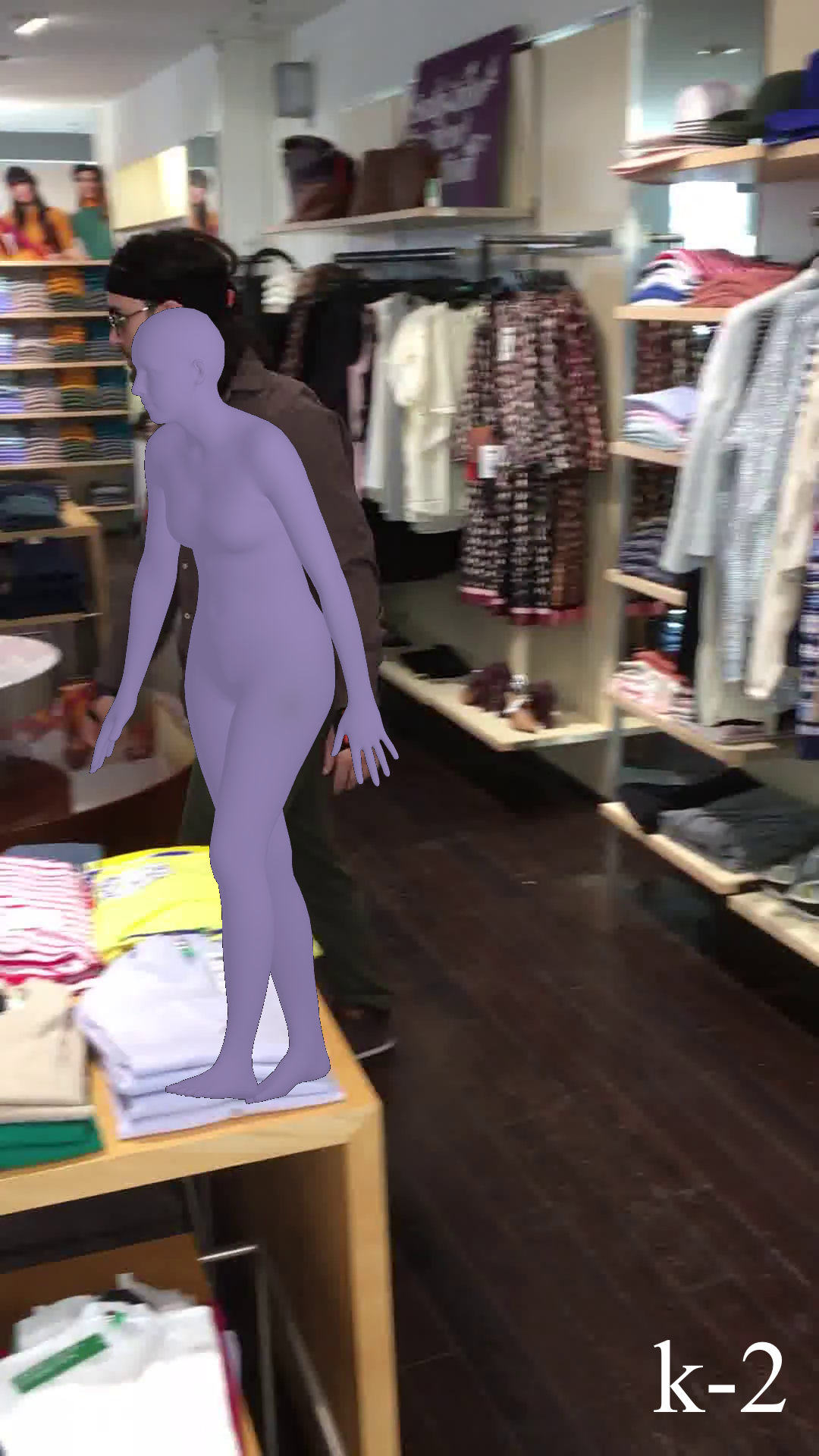}}
\quad
{\includegraphics[width=0.20\textwidth]{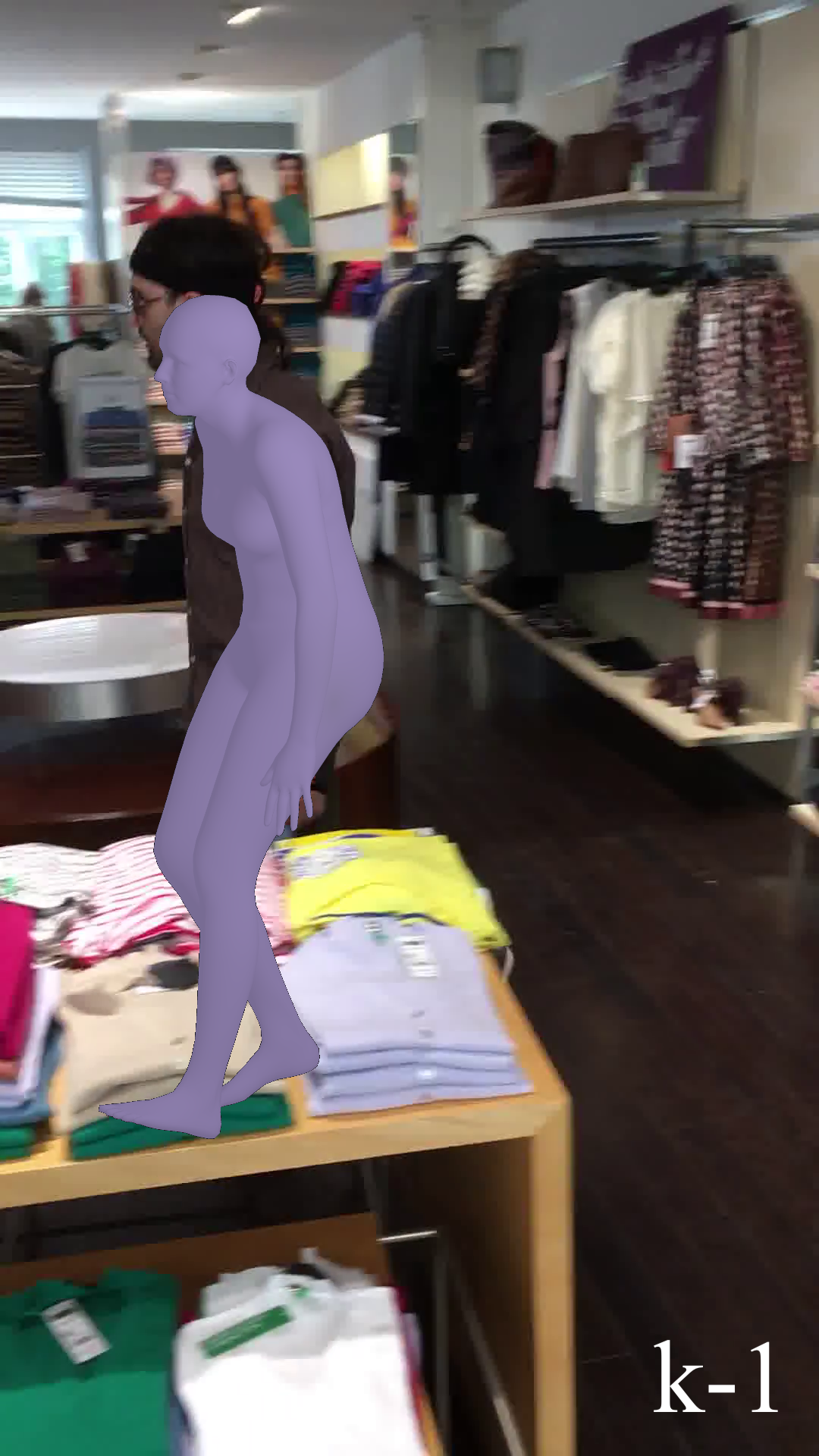}}
\quad
{\includegraphics[width=0.20\textwidth]{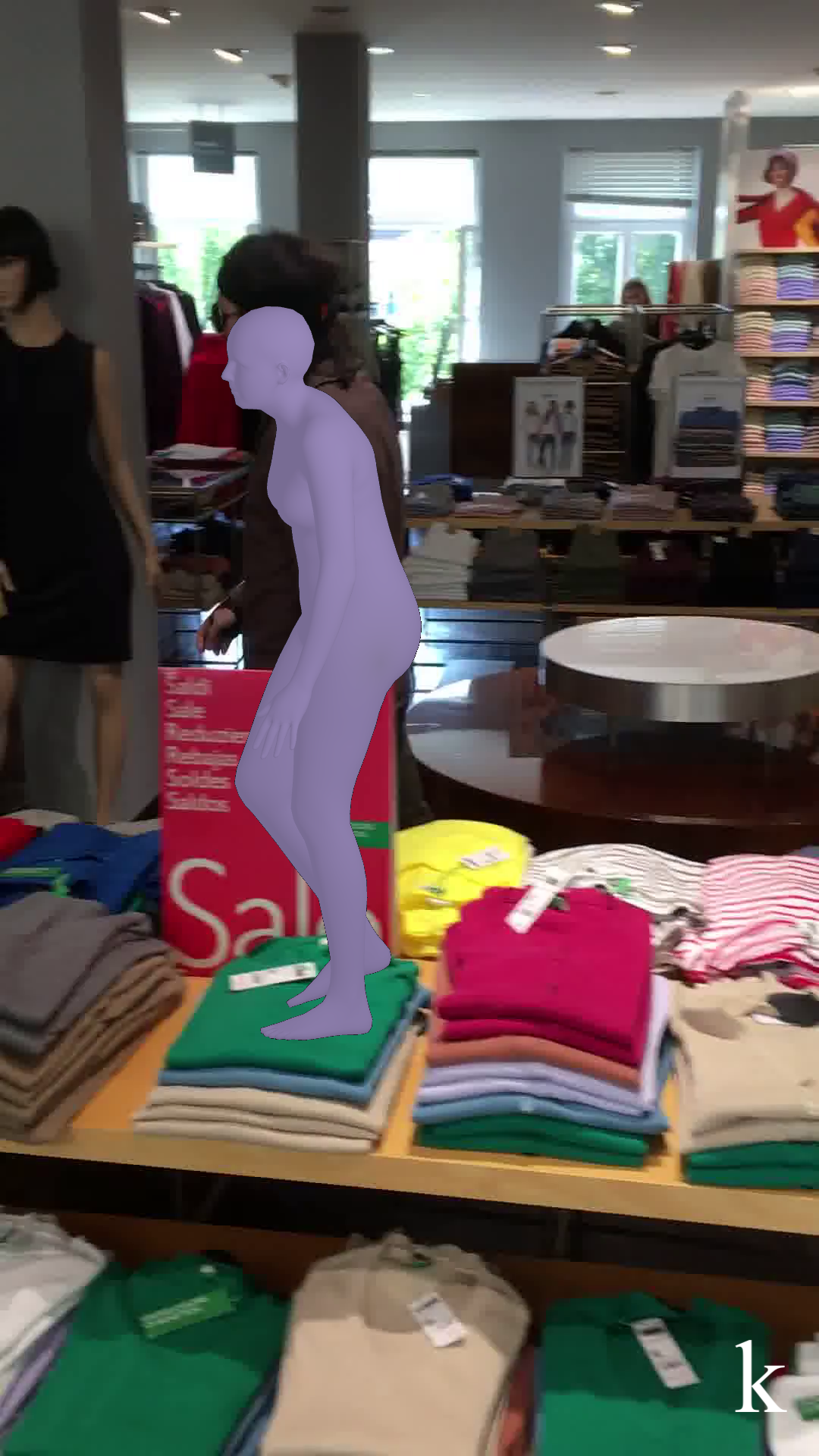}}
{\includegraphics[height=180pt]{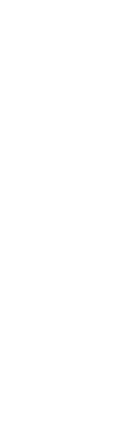}}
\end{figure*}
\begin{figure*}[ht!]
\centering
\includegraphics[width=0.20\textwidth]{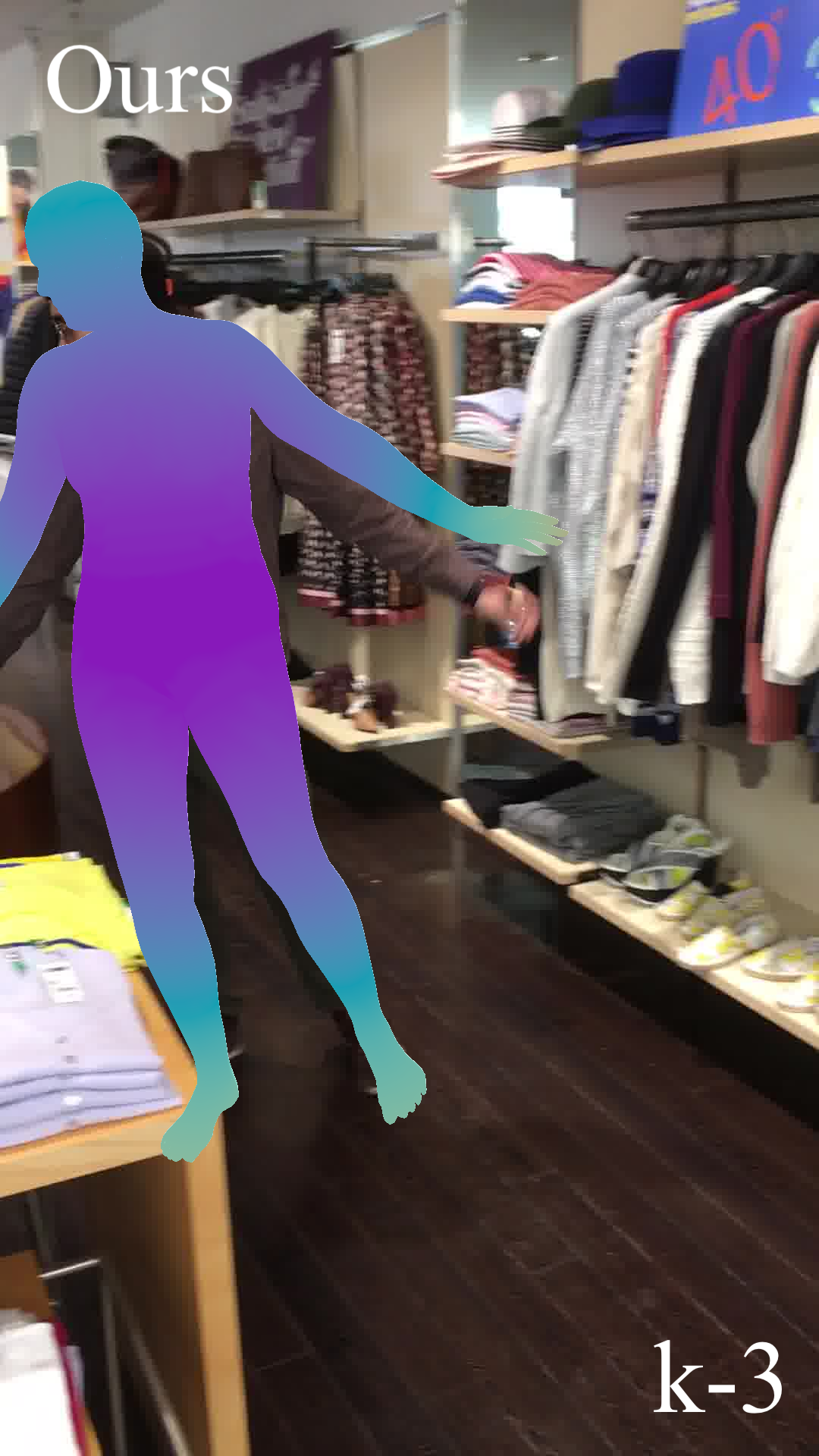}
\quad
{\includegraphics[width=0.20\textwidth]{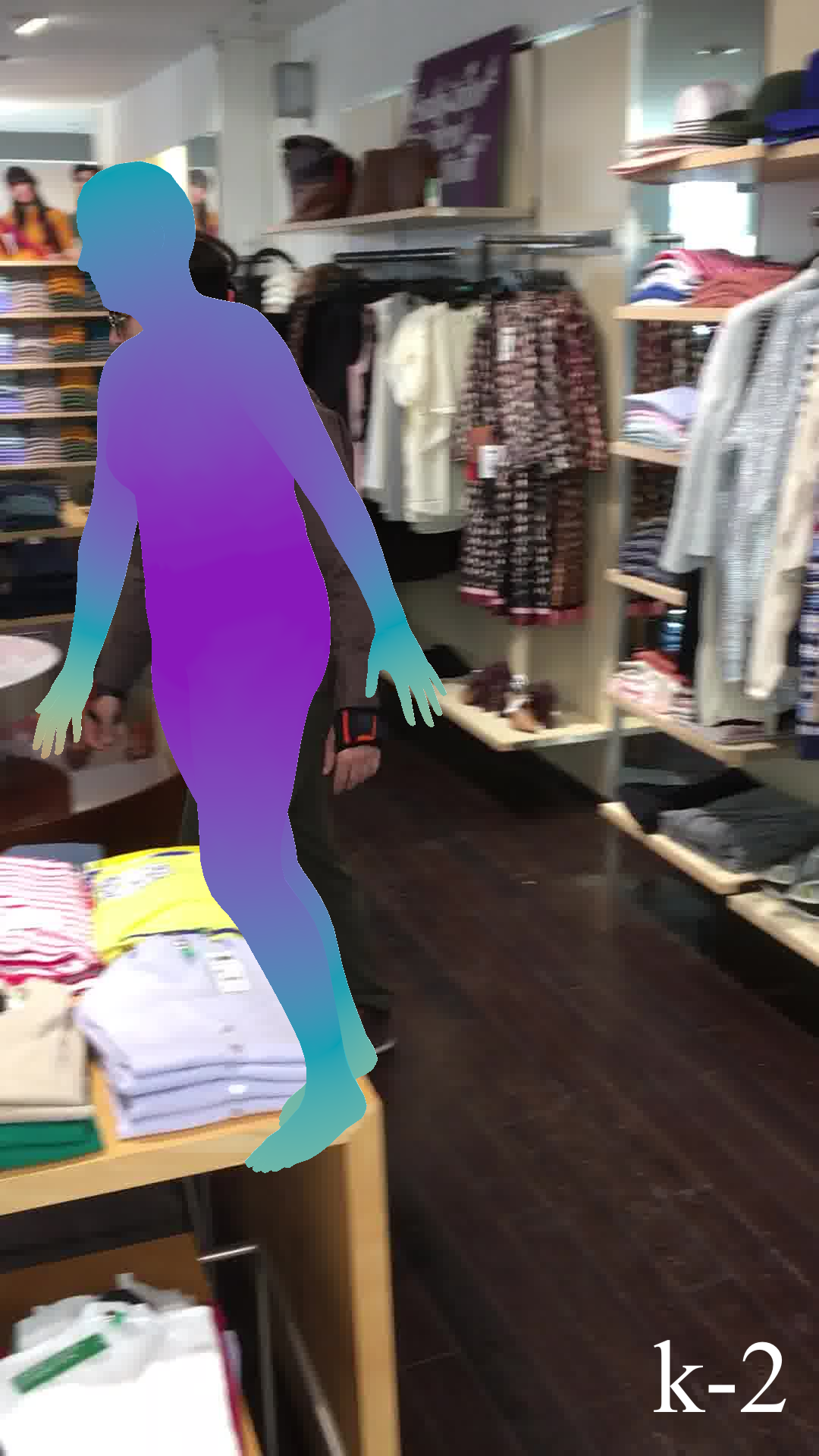}}
\quad
{\includegraphics[width=0.20\textwidth]{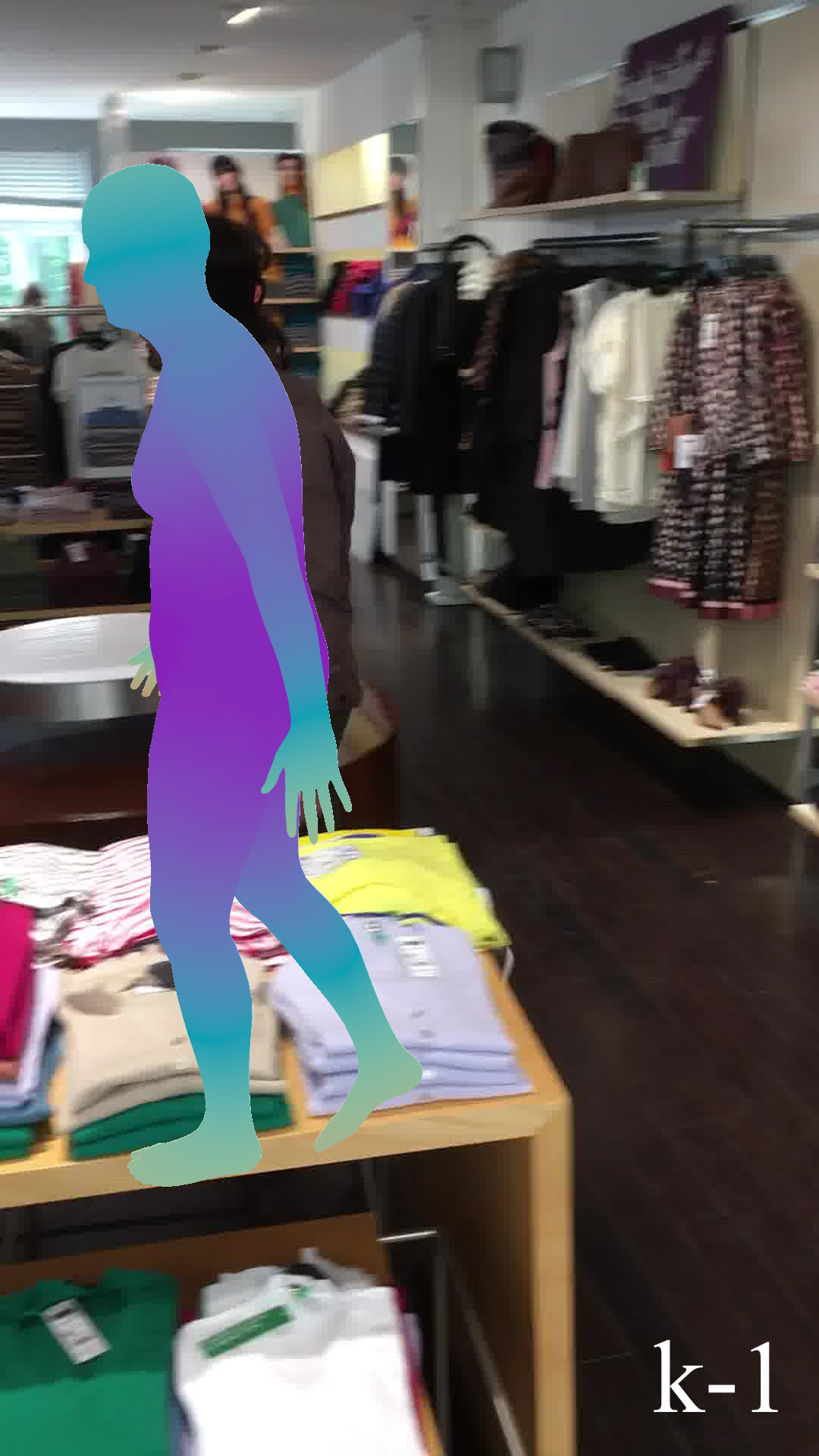}}
\quad
{\includegraphics[width=0.20\textwidth]{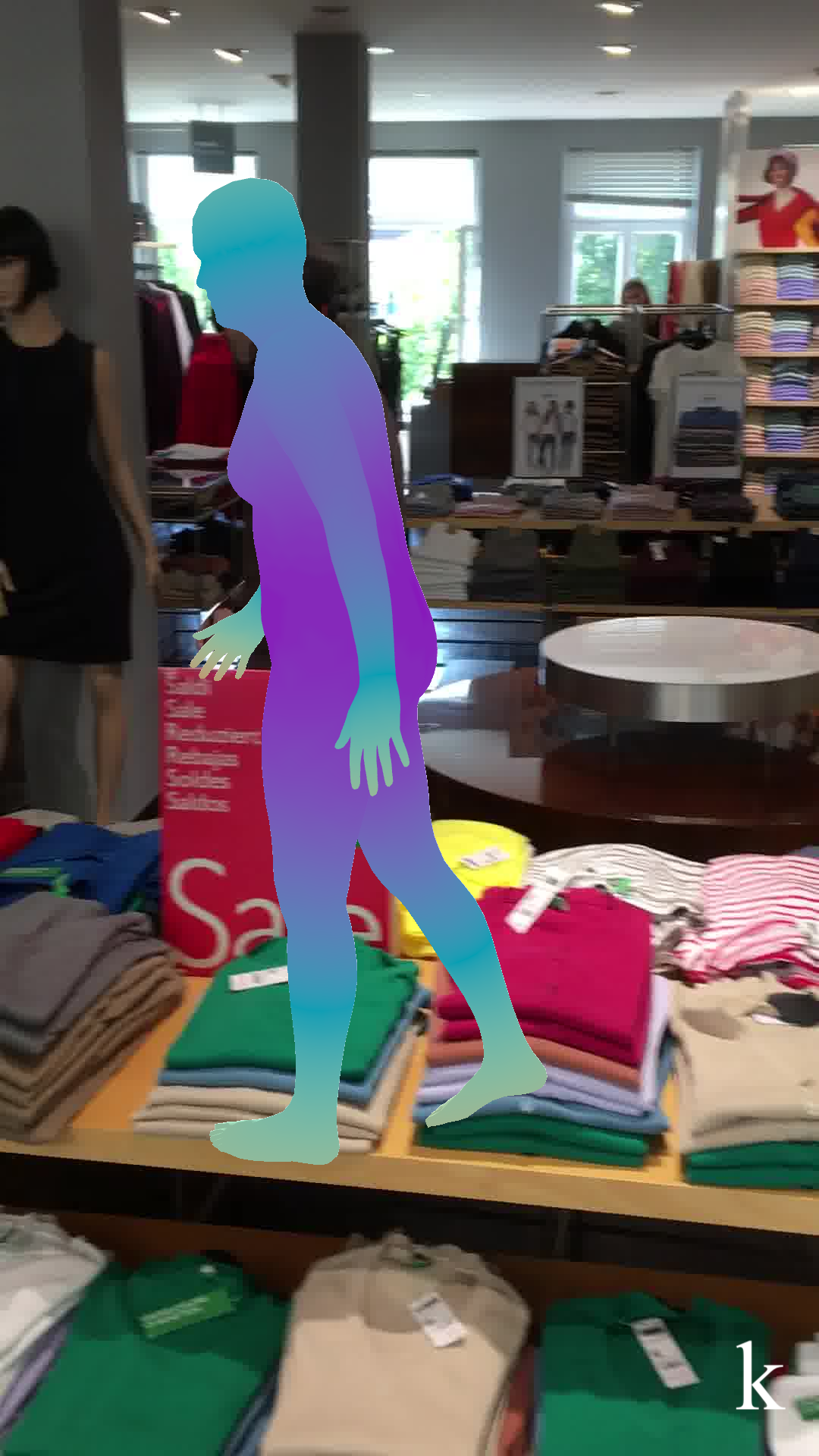}}
{\includegraphics[height=180pt]{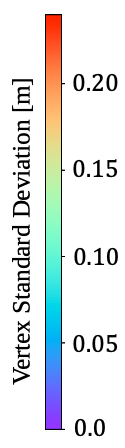}}
\caption{Qualitative comparison of the human body tracking with VIBE~\cite{kocabas2019vibe} (upper) and our method (lower) in the presence of severe occlusions. Our method yields a body state distribution with a stable body posture and root pose prediction, as well as a reasonable uncertainty in the occluded areas while tracking the body over the whole sequence. Meanwhile, VIBE only predicts a deterministic human body mesh, loses track under occlusion and consequently estimates a wrong body posture.}
\label{fig:qualitative}
\end{figure*}

As qualitatively demonstrated in Fig.~\ref{fig:qualitative}, our method is able to deal with severely occluded areas by leveraging its motion-based prior.
Both presented models fail to identify the correct body shape.
VIBE \cite{kocabas2019vibe} loses track under part occlusion while confidently predicting the wrong body posture and shape.
In contrast, our method is able to keep tracking the body with a stable posture and root pose estimation while additionally estimating a reasonable uncertainty distribution in the occluded and misaligned area.

\section{Conclusion}
\label{sec:conclusion}
An accurate, uncertainty-aware 3D human pose estimation method is key to every human-robot interaction task.
In this work, we presented GloPro, the first method that is able to predict the full 3D body mesh uncertainty distribution including body shape and root pose.
In the underlying approach, GloPro efficiently fuses visual predictions with a learned motion model in real-time for multiple body meshes. 
We demonstrate that this architecture facilitates accurate body pose estimates under severe short-term occlusions while providing a relatively consistent uncertainty.
Also, we showed how an external camera pose estimation can be effectively induced into the system to further robustify the prediction.
Future work might extend our probabilistic framework to multi-modal body state distributions such as a mixture of Gaussians, to enable improved probabilistic modeling even in cases of long-term occlusions.  

\section*{Acknowledgements}
We thank Prof.\ Dr.\ Kathrin Dörfler and Lidia Atanasova for the valuable discussions and many practical insights.

\bibliographystyle{IEEEtran}
\bibliography{references}

\end{document}